%% file: acl_latex.tex
\pdfoutput=1

\documentclass[11pt]{article}

\usepackage[final]{acl}

\usepackage{times}
\usepackage{latexsym}
\usepackage{stmaryrd}
\usepackage{mathtools}
\usepackage{multirow}
\usepackage{pifont}
\usepackage[table]{xcolor}
\usepackage[T1]{fontenc}

\usepackage{color}

\newcommand{\tower}{\textsc{Tower}}
\newcommand{\spire}{\textsc{Spire}}
\newcommand{\spbase}{\textsc{SpireBase}}
\newcommand{\spinstruct}{\textsc{SpireFull}}
\newcommand{\itonly}{\textsc{TowerFull}}
\newcommand{\noblocks}{\textsc{SpireNoBlocks}}
\newcommand{\nopseudo}{\textsc{SpireNoPseudo}}

\newcommand{\langp}[2]{#1$\rightarrow$#2}
\newcommand{\langpb}[2]{#1$\leftrightarrow$#2}

\usepackage[utf8]{inputenc}

\usepackage{microtype}

\usepackage{inconsolata}

\usepackage{graphicx}

\usepackage{booktabs}
\usepackage{arydshln}
\title{From \tower{} to \spire{}: Adding the Speech Modality to a Translation-Specialist LLM}

  \author{
 \textbf{Kshitij Ambilduke\thanks{Equal contribution. $^\P$Equal contribution.}\thanks{Work begun during an internship at Instituto de Telecomunicações.}\textsuperscript{1}},
 \textbf{Ben Peters$^*$\thanks{Work done while at Instituto de Telecomunicações.}\textsuperscript{2}},
 \textbf{Sonal Sannigrahi$^*$\textsuperscript{3,4}},
 \textbf{Anil Keshwani$^\dagger$\textsuperscript{5}},
\\
 \textbf{Tsz Kin Lam\textsuperscript{6}},
 \textbf{Bruno Martins\textsuperscript{2,4}},
 \textbf{André F.T. Martins$^\P$\thanks{Work done while at Unbabel.}\textsuperscript{3,4,7,8}},
  \textbf{Marcely Zanon Boito\textsuperscript{$\P$9}}
\\
\\
 \textsuperscript{1}ENS Paris-Saclay
 \textsuperscript{2}INESC-ID
 \textsuperscript{3}Instituto de Telecomunicações\\
 \textsuperscript{4}Instituto Superior Técnico, Universidade de Lisboa
 \textsuperscript{5}Sapienza University of Rome\\
 \textsuperscript{6}University of Edinburgh
 \textsuperscript{7}ELLIS Unit Lisbon
 \textsuperscript{8}TransPerfect
 \textsuperscript{9}NAVER LABS Europe
\\
 \small{
   \textbf{Correspondence:} \href{mailto:benzurdopeters@gmail.com}{benzurdopeters@gmail.com}
 }
}

\begin{document}
\maketitle
\begin{abstract}
We introduce \spire{}, a speech-augmented language model (LM) capable of both translating and transcribing speech input from English into 10 other languages as well as translating text input in both language directions. 
\spire{} integrates the speech modality into an existing multilingual LM via speech discretization and continued pre-training using only $42.5$K hours of speech. 
In particular, we adopt the pretraining framework of multilingual LMs and treat discretized speech input as an additional \textit{translation language}. This approach not only equips the model with speech capabilities, but also preserves its strong text-based 
performance. We achieve this using significantly less data than existing speech LMs, demonstrating that discretized speech input integration as an additional language is feasible during LM adaptation. We make our code and models available to the community.
\end{abstract}

\input{latex/sections/1_introduction}
\input{latex/sections/2_background}

\input{latex/sections/3_methodology}
\input{latex/sections/4_results}
\input{latex/sections/6_conclusion}

\bibliography{custom,anthology}

\input{latex/sections/6_appendix}

\end{document}

%% file: latex/sections/1_introduction.tex
\section{Introduction}

Large language models (LLMs) have demonstrated remarkable success on various text-based natural language processing tasks~\cite{achiam2023gpt, touvron2023llama, yang2024qwen2, alves2024tower,martins2024eurollm}, motivating research into extending them to other modalities. This has led 
to the development of multimodal LMs capable of processing speech, audio, images, and video~\cite{team2023gemini, driess2023palm, rubenstein2023audiopalm,liu_visual_2023, tangsalmonn,defossez2024moshi,hu2024wavllm, huang2024audiogpt, nguyen2024spirit}. However, the integration of new modalities often comes at the cost of existing capabilities~\cite{zhai2024investigating}.

For speech-LLM integration, a simple approach is to link the output of an automatic speech recognition~(ASR) system to a text-only LLM~\cite{huang2024audiogpt}. This solution, however, is prone to error propagation and depends largely on individual model quality. More popular are solutions that investigate equipping LLMs natively with speech processing capabilities through modality projection~\cite{shu2023llasm, radhakrishnan-etal-2023-whispering, wu2023decoder, tangsalmonn, xue2024ideal,hu2024wavllm}. 
Typically, a speech foundation model generates speech representations that are mapped to the embedding space of the LLM, following which the model is then fine-tuned along with a projector on speech-to-text tasks to equip the LLM with speech processing capabilities. In this setting, key challenges include prompt overfitting and high training costs, as tuning these multimodal LLMs requires the adaptation of the speech projector module on vast amounts of raw speech data~\cite{tangsalmonn, hu2024wavllm}.

An alternative approach for 
integrating speech into a text-only LLM is to use \textit{speech discretization}, where continuous speech features are transformed prior to training into sequences of ``discrete speech units''~(DSUs), which can be processed similarly to text~\cite{chou2023toward, zhang-etal-2023-speechgpt, rubenstein2023audiopalm, chang2024exploring, defossez2024moshi, trinh2024discrete, maiti2024voxtlm, nguyen2024spirit}. 
This approach simplifies training by eliminating the need for additional parameters beyond extended embedding matrices. 
Finally, while both projector-based and discretization-based solutions have shown promising results on text-to-speech and speech-to-text tasks, their development has prioritized speech-centric tasks at the expense of textual performance. Furthermore, limited research has focused on integrating speech while preserving the LLM's original capabilities in textual tasks~\cite{chou-etal-2023-toward,huang2024audiogpt}.

\input{latex/figures/pipeline}

In this work we present \spire{}, a speech-augmented LLM built from the open-weight multilingual model \tower{}~\cite{alves2024tower}. \spire{} can perform English ASR and from-English speech translation~(ST) while maintaining \tower{}'s strong performance on machine translation~(MT) across all 10 languages\footnote{en, de, fr, nl, it, es, pt, ko, ru, zh} supported by \tower{}. 
\spire{} encodes speech via HuBERT-based~\cite{hsu2021hubert} k-means clustering, as in previous work~\cite{zhang-etal-2023-speechgpt,rubenstein2023audiopalm,chang2024exploring}. We perform training in two stages: Continued Pre-Training~(CPT) and Instruction Tuning~(IT). For the CPT stage, we use a mixture of ASR data and a small fraction of \tower{}'s text CPT data. For IT, we leverage \tower{}'s task-specific MT data, as well as additional English ASR and ST data. \spire{} is trained using only \textbf{42.5K} hours of speech, differing from the large scale of data used by existing models \cite{radford2023robust, nguyen2024spirit, Qwen2-Audio}. Figure~\ref{fig:model} illustrates our training process. We make the following contributions:

\begin{itemize}
    \item We present a pipeline for integrating speech as an additional modality into an existing LLM, enabling it to transcribe and translate English speech while preserving its original text capabilities across 10 languages;
    \item We analyze speech integration at two stages, namely CPT and IT, demonstrating the necessity of both stages to achieve optimal performance across both modalities; 
    \item We make our models, datasets, and scripts available to the community.\footnote{\url{https://huggingface.co/collections/utter-project/spire-67d4253d6af8d6a0308527e0}} 
\end{itemize}

%% file: latex/figures/pipeline.tex
\begin{figure*}
    \centering
    \includegraphics[width=0.9\linewidth]{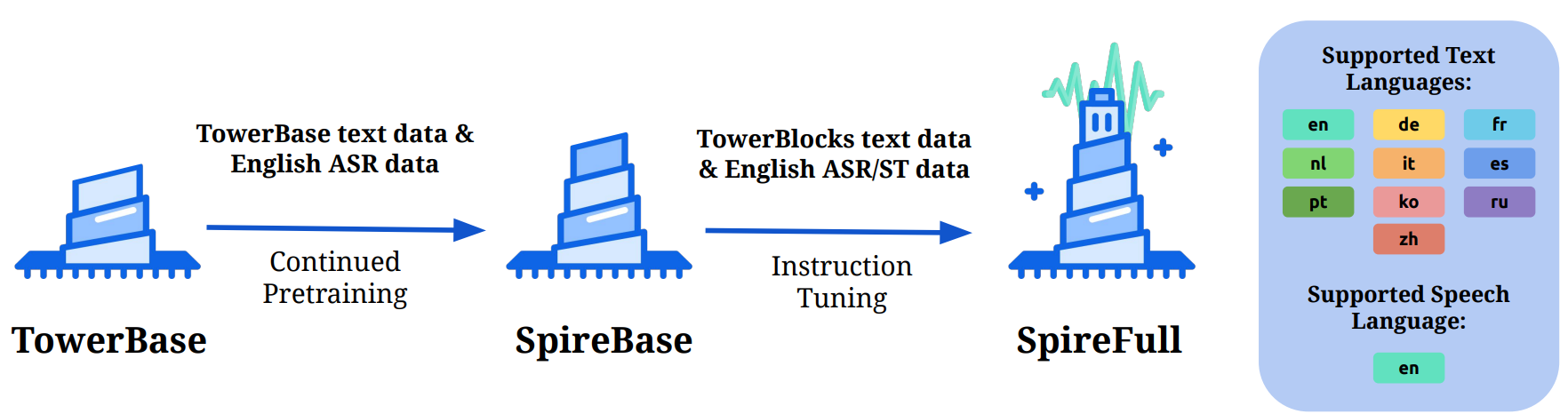}

    \caption{Illustration of the model training approach for \spbase{} and \spinstruct{}.
    }
    \label{fig:model}
\end{figure*}

%% file: latex/sections/2_background.tex
\section{Related Work}

\paragraph{Speech-to-Text Models} An increasing number of studies have explored integrating speech into LLMs~\cite{zhang-etal-2023-speechgpt, rubenstein2023audiopalm,hassid2024textually}. For discrete speech input, \citet{hassid2024textually} demonstrate the benefits of initializing a speech LLM from a text-based LLM.  
SpeechGPT~\cite{zhang-etal-2023-speechgpt} 
applies IT on speech-to-text 
ASR, text-to-speech~(TTS), and text-based question answering. AudioPALM~\cite{rubenstein2023audiopalm} 
is trained in a multi-task fashion, similarly to SpeechGPT, but on multilingual input.  
Recently, VoxtLM~\cite{maiti2024voxtlm} was trained jointly on DSUs 
and text data for ASR, TTS, and open-ended speech/text generation. 
Our work is most similar to SpiritLM \citep{nguyen2024spirit}, which adapts an LLM with an interleaved mixture of DSU and text data, which requires an expensive DSU-to-transcript step to create. 
In contrast, we adopt a more cost-effective input representation that can be extended to any language, regardless of the availability of a speech aligner.
Our focus is on successfully incorporating speech input while preserving the original competence of the model, so that the resulting model can successfully perform both speech-to-text and text-only tasks. None of the aforementioned models are trained to preserve the original model's performance in text tasks.

\paragraph{Adapting LLMs} Previous approaches involve training from scratch with task- and domain-specific data~\cite{singhal2023large, lewkowycz2022solving}, performing CPT with a diverse training data mix designed to broadly extend the model's knowledge~\cite{wu2023bloomberggpt}, or IT 
on 
use-case-specific data~\cite{chen2023meditron}. Recent work has explored combining the latter two approaches~\cite{xuparadigm,alves2024tower, weifinetuned,roziere2023code}. In our approach to integrating DSUs into \tower{}, we take inspiration from \citet{alves2024tower} in adopting a two-step CPT+IT process.
Our work differs in that we focus on adding the speech modality, whereas \citet{alves2024tower} focused on increasing the multilingual capabilities of an LLM. 

\paragraph{Continuous and Discrete Speech Representations} 
Self-supervised speech representation models produce contextualized high-dimensional speech vectors directly from raw audio~\cite{hsu2021hubert,baevski2020wav2vec,chen2022wavlm}, largely outperforming statistical speech features on
downstream tasks~\cite{yang2021superb}. These continuous representations can be used to derive DSUs that capture both linguistic content and prosody through clustering~\cite{borsos2023audiolm,kharitonov-etal-2022-text}. DSUs provide better alignment with textual data, facilitating the transfer of successful training settings from the text domain~\cite{cui2024recent}. Building on~\citet{lakhotia-etal-2021-generative}, which demonstrated that HuBERT~\cite{hsu2021hubert}
is a powerful feature extractor, several studies have adopted this approach, incorporating a k-means clustering step for discretization~\cite{zhang-etal-2023-speechgpt,rubenstein2023audiopalm,lam2024compact,chang2024exploring,nguyen2024spirit}. \citet{xucomparing} study the optimal settings to obtain DSUs in terms of cluster size and feature extraction layer. We use their findings to inform our initial choices.

%% file: latex/sections/3_methodology.tex
\section{\spire{}: A Speech-to-Text LLM}
We introduce \spire{}, whose goal is to equip an LLM with speech capabilities while preserving its preexisting text capabilities.
As our base LLM we choose \tower{}~\cite{alves2024tower}, which was developed from Llama-2~\cite{touvron2023llama} with a two-step approach: CPT on a mixture of monolingual and parallel data (\tower{\textsc{Base}}), followed by IT on translation-related tasks (\tower{\textsc{Instruct}}).
We use an approach similar to \tower{} to extend the model to the speech modality.
First, we perform CPT with a combination of text-only and aligned speech-to-text datasets, followed by IT using both text-only general-purpose and task-specific data curated in \textsc{TowerBlocks},\footnote{\scriptsize\url{https://huggingface.co/datasets/Unbabel/TowerBlocks-v0.2}} alongside task-specific speech-to-text datasets. 

We choose \tower{} in particular due to its competitive performance compared to other open alternatives. \tower{}-based models were among the best participating systems in the WMT24 general translation task \cite{wmt24-findings}. \tower{}'s usage of open source data during the CPT phase along with the release of the \tower{\textsc{Blocks}} dataset, used in the IT phase, further motivates our choice.

\subsection{Speech Discretization}

To easily transfer the training setup 
of \tower{}, we use DSUs as opposed to an auxiliary speech encoder. For all speech datasets that were used, 
we follow recent discretization methodology~\cite{zhang-etal-2023-speechgpt,rubenstein2023audiopalm,chang2024exploring} to produce DSUs by first extracting continuous speech representations for our speech data from the 22nd layer of an HuBERT-large model, trained on 60K hours of English speech \cite{hsu2021hubert}, and then using k-means clustering~($K=5000$) to produce centroids that are used to convert our continuous speech representation into a discrete sequence of cluster IDs.\footnote{Optimizing the layer selection for feature extraction is a complex research problem~\cite{pasad2023comparative,mousavi2024should}. In this work we follow the insights from \citet{gow-smith-etal-2023-naver} and \citet{xucomparing}.} We train our k-means model on a collection of 235K audio files~(approximately 720 hours), drawn from three speech corpora: CoVoST-2~\citep{wang2021covost}, VoxPopuli~\cite{wang-etal-2021-voxpopuli}, and Multilingual LibriSpeech~\citep[MLS; ][]{pratap20_interspeech}.
The CoVoST subset consists of 62K 
audio files from 10,049 speakers, with a maximum of 8 audio files per speaker. The VoxPopuli subset includes 65K 
audio files from 639 speakers, capped at 
250 audio files per speaker. Finally, the MLS subset contains 107K audio files from 5,490 speakers.

\subsection{\spbase{}}

The first CPT stage, yielding \spbase{}, is trained from \tower{\textsc{Base-7B}}\footnote{We used \tower{}-7B models instead of the 13B or 70B versions due to its lower compute requirements.} using both text-only and aligned speech-to-text datasets. Following previous work, we include a fraction of \tower{}'s original training data to preserve its existing performance~\cite{scialom-etal-2022-fine, de2019episodic}.

\subsubsection{Data}\label{sec:data:cp}

We use a mixture of monolingual and parallel text in Chinese~(zh), Dutch~(nl), English~(en), French~(fr), German~(de), Italian~(it), Korean~(ko), Portuguese~(pt), Russian~(ru), and Spanish~(es), that was sourced from the \tower{} training data, as well as English ASR data sourced from popular open-source ASR datasets, as reported in Table~\ref{tab:speechstatistics}. Both speech and text data are downsampled to create a 6B token data mixture~(5B speech; 1B text), measured by the model tokenizer.\footnote{Preliminary experiments on the data mixture led to this particular choice.}
Note that the 5B speech tokens include both DSUs (4.4B tokens) and their text transcriptions (0.6B tokens).

\paragraph{Text Data} The monolingual text data split corresponds to data from mC4~\cite{2019t5}, a multilingual web-crawled corpus which we uniformly sample from across all languages. The parallel data split includes uniformly sampled instances to and from English (\langpb{en}{xx}) for the 10 languages, sourced from various public sources. Further details can be found in \citet{alves2024tower}.

\paragraph{Speech Data} We collect 35K hours of speech data from SPGI Speech~\cite{o2021spgispeech}, GigaSpeech~\cite{chen21o_interspeech}, MLS, and VoxPopuli. We normalize as described in Appendix~\ref{appendix:datanorm}. 

\subsubsection{CPT Setup}

We train \spbase{} using MegatronLLM~\cite{epfmgtrn} on 8 A100-80GB GPUs for 6 days. We use the same hyperparameters as \tower{}, except for the effective batch size, which in our case is 2304.
To incorporate the DSUs in the CPT stage, we extend the model's original vocabulary by 5000 types, \textit{e.g.},~\texttt{<extra\_id\_x>}.
This allows us to have a vocabulary that can encode both text in subword units and speech in DSUs. 
For the extended vocabulary, we initialize new embeddings from a multivariate Gaussian distribution. 
The mean of this distribution is set to the average of the original embeddings, while the covariance is derived from the empirical covariance of the original embeddings, scaled by a factor of $1 \times 10^{-5}$ \cite{hewitt2021initializing}. 

\subsection{\spinstruct{}}
\spinstruct{} is obtained by instruction tuning \spbase{} on task-specific text and speech data.

\subsubsection{Data}\label{sec:data:it}
We use a mixture of text and speech instructions for ASR, MT, and ST. The prompt formats used during training are shown in Appendix~\ref{appendix:prompts}.

\paragraph{Text Data} We use \textsc{TowerBlocks}~\cite{alves2024tower}, which includes high quality translation bi-texts between English and the other languages supported by \tower{}. It also includes instructions for 
the translation-related tasks of named entity recognition~(NER) and automatic post-editing~(APE).

\paragraph{ASR Data} We use 0.8K hours of ASR data from CommonVoice 18~\citep[CV; ][]{ardila2019common}, downsampling as described in Appendix~\ref{appendix:datanorm}.

\paragraph{ST Data} 
In our IT set, we use 842 hours of speech across 
three 
ST training sets: FLEURS (all nine language pairs; we filter out examples whose transcriptions overlap with the FLORES devtest set), Europarl-ST~\cite{iranzo2020europarl}~(en $\shortrightarrow$ \{de, es, fr, it, nl, pt\}), and CoVoST-2 (\langp{en}{zh}).
Since this amounts to far less data for ST than ASR, and since \langp{en}{\{ko, ru\}} have only examples from the tiny FLEURS set, we augment our speech collection with \textbf{pseudo-labeled} data, which has been effective for other ST systems~\citep{barrault2023seamlessm4t}.
We select 300k ASR examples each from CV, SPGI, and GigaSpeech and translate them to all nine target languages using \texttt{TowerInstruct-13B}.\footnote{\scriptsize{\url{https://huggingface.co/Unbabel/TowerInstruct-13B-v0.1}}}
We then filter examples whose transcript-translation combination has a COMET-QE\footnote{\scriptsize{\url{https://huggingface.co/Unbabel/wmt22-cometkiwi-da}}}~\citep{rei-etal-2022-cometkiwi} score under 85.
Finally, for each language pair, we 
sample 60K examples to be used in direct ST prompts and another 60K to be used in multi-turn prompts. 
This results in
180K direct ST prompts and 180K multi-turn prompts for each language pair.\footnote{Due to our aggressive filtering, we were left with slightly fewer examples for en $\shortrightarrow$ zh.} The prompt formats are shown in Appendix~\ref{appendix:prompts}.

\input{latex/tables/speech_statistics}

\subsubsection{IT Training Setup}

We use the \texttt{chatml} template \cite{openai2023chatml} to format our instructions in dialogue form. We train models using Axolotl\footnote{\scriptsize\url{https://github.com/axolotl-ai-cloud/axolotl}} on 4 H100-80GB GPUs for 2.7 days. We use a learning rate of $7 \times 10^{-6}$ and a cosine scheduler with 100 warm-up steps. We train for 4 epochs with an effective batch size of 576 and a weight decay of 0.01. We impose a maximum sequence length of 4096 and use the AdamW optimizer~\cite{loshchilov2019decoupled}. Other hyperparameters are derived from \tower{\textsc{Instruct}}~\citep{alves2024tower}.

%% file: latex/tables/speech_statistics.tex


\begin{table}
\resizebox{\columnwidth}{!}{\begin{tabular}{cccrr}
\toprule
\textbf{Dataset} & \textbf{Task} & \textbf{Phase} & \textbf{\# DSUs} & \textbf{\# Hours} \\ \midrule

SPGI Speech & ASR & CPT & 645M & 5.1K \\
Gigaspeech & ASR & CPT & 1.2B & 9.9K\\
MLS & ASR & CPT & 2.4B & 19.2K\\
VoxPopuli & ASR & CPT & 69M & 0.5K\\
CV & ASR & IT & 105M & 0.8K\\
Europarl-ST & ST & IT & 122M & 1.0K\\
FLEURS & ST & IT & 11M & 0.09K\\
CoVoST-2 & ST & IT & 12M & 0.09K\\
SPGI Speech & Pseudo-ST & IT & 350M & 2.8K\\
GigaSpeech & Pseudo-ST & IT & 161M & 1.3K\\
CV & Pseudo-ST & IT & 212M & 1.7K\\
\bottomrule
\end{tabular}}
\caption{Statistics for speech training data. Hours are approximated from the number of deduplicated DSUs.}
\label{tab:speechstatistics}
\end{table}

%% file: latex/sections/4_results.tex
\section{Experiments}\label{sec:experiments}

We evaluate our models across three tasks: ASR, MT, and ST. First, we present our results for ASR~(\S\ref{sec:asreval}), confirming the new capabilities \spire{} has in the speech domain.
We then present MT results~(\S\ref{sec:texteval}), demonstrating that the speech performance does not come at the expense of the original model's MT performance. Finally, we present results for ST~(\S\ref{sec:asteval}) to investigate model performance on a task that requires both 
ASR and MT capabilities. 

\paragraph{Evaluation Setup}
Across models and tasks, we perform inference with greedy decoding with a maximum of 256 generated tokens.
For the \tower{} and \spire{} models, we decode with \texttt{vllm}~\cite{kwon2023efficient}.
However, since \texttt{vllm} does not support all of our baselines, we use alternative libraries~\citep[\texttt{transformers};][]{wolf2019huggingface} where necessary.
Unless specified otherwise, we use zero-shot prompts for all models and tasks.

\input{latex/sections/4_asr_results}
\input{latex/sections/4_mt_results}
\input{latex/sections/4_st_results}

%% file: latex/sections/4_asr_results.tex
\subsection{ASR}\label{sec:asreval}

\paragraph{Datasets and Metrics}
We evaluate ASR performance across multiple test sets, in order to cover a variety of recording styles: LibriSpeech~(LS) test-clean and test-other~\cite{panayotov2015librispeech}, FLEURS~\cite{conneau2023fleurs}, and VoxPopuli.\footnote{For CPT models, LS is an in-domain evaluation because its training set is part of MLS.} We report the Word Error Rate (WER) between the hypotheses and gold transcripts, after Whisper normalization~\citep{radford2023robust}.

\paragraph{Baselines} We include the following models:

\begin{itemize}
    \item \textbf{Whisper}~\citep{radford2023robust} is an encoder-decoder transformer trained on over 5 million hours of labeled data that performs multilingual ASR and to-English ST. We report results for Whisper-base (74M parameters) and Whisper-large-v3 (1.5B parameters).
    \item \textbf{SeamlessM4T}~\citep{barrault2023seamlessm4t} is an encoder-decoder transformer trained on 406K hours of speech that performs ASR, ST and MT across 100 languages. We report results for SeamlessM4T-large-v2 (2.3B parameters).
    \item \textbf{SALMONN}~\cite{tangsalmonn} integrates a pre-trained text LLM with separate speech and audio encoders into a single multimodal model.\footnote{SALMONN uses 4400 hours of speech/audio data in the IT phase but does not specify the large amount of pre-training ASR and audio captioning data used.} SALMONN uses a LoRA adapter~\cite{hulora} to align the spaces. 

    \item \textbf{Qwen2-Audio}~\cite{Qwen2-Audio} integrates audio into Qwen-7B \citep{qwenlm} using a specialized encoder that is initialized from Whisper large-v3. The resulting model is pretrained on $\sim$520K hours of data spanning speech, sound, and music.
    
    
    \item \textbf{SpiritLM}~\cite{nguyen2024spirit} is a decoder-only model, trained from Llama-2 on 307B tokens of text, 458K hours of unlabeled speech, and 111K hours of labeled speech. As in \spire{}, it uses HuBERT DSUs. 

    \item \textbf{HuBERT-large+CTC} is a CTC-based ASR model trained using the same speech representation model we use for DSU generation, and using the same ASR data from the IT stage~(Section~\ref{sec:data:it}).\footnote{The hyperparameters are described in Appendix~\ref{appendix:model:asrmodel}.} Unlike \spire{}, this model has access to a very powerful speech representation backbone. However, it lacks strong language modeling capabilities.
\end{itemize}

\input{latex/tables/ASR_table}
\paragraph{Results} 
Our results are presented in Table~\ref{tab:results_asr}.
\spinstruct{}'s performance demonstrates that performing both the CPT and IT stages is an effective strategy to give speech capabilities to a text LLM. On the other hand, \spbase{} does not consistently show reasonable speech performance; however, on FLEURS and VoxPopuli we obtain somewhat strong results in the 
zero-shot settings, which is surprising given that non-instruction-tuned models often struggle to work out-of-domain without in-context learning examples.\footnote{We also tried prompting \spbase{} with few-shot examples, but the results were much worse, possibly because the length of the DSU sequences led to in-context examples that were too long for the model to handle effectively.}

Although \spinstruct{} does not match the performance of SeamlessM4T, Whisper-large-v3, SALMONN, or Qwen2-Audio, these were trained on far more speech data than our models (around 10x for Qwen2-Audio and SeamlessM4T).
Given this training data gap, it is notable that \spinstruct{} \textit{does} outperform Whisper-base on LS and FLEURS, and SpiritLM on all benchmarks SpiritLM reports, at a fraction of the speech data.

\spinstruct{} also outperforms the HuBERT-large+CTC baseline on three out of four datasets. This is an impressive result given that the CTC model has access to continuous features, which \spinstruct{} lacks. We believe this demonstrates that our compressed discrete representations capture the speech signal well enough to support speech-to-text tasks.

%% file: latex/tables/ASR_table.tex
\begin{table}[t]
    \centering
    \small
\resizebox{\columnwidth}{!}{\begin{tabular}{lrrrr}
\toprule
 {} & \multicolumn{2}{c}{\textbf{LibriSpeech}} & \multirow{2}{*}{\textbf{FLEURS}} & \multirow{2}{*}{\textbf{VoxPopuli}}\\
  {} & Clean & Other &  & \\
 \midrule
 Whisper-base & 5.0 & 11.9 & 12.1 & 9.8\\
 Whisper-large-v3 & 1.8 & \textbf{3.7} & \textbf{5.8} & 9.2\\
 SeamlessM4T & 2.6 & 4.9 & 8.1 & 7.5 \\
 SALMONN & 2.4 & 5.3 & 9.3 & 8.9\\
 Qwen2-Audio & \textbf{1.6} & 3.9 & 6.6 & \textbf{6.5}\\
 SpiritLM & \textit{6.0}* & \textit{11.0}* & - & - \\
  HuBERT-large+CTC & 4.3 & 7.6 & 11.4 & 14.7\\ 
\midrule
\textit{Our models} \\
\hdashline 
\rowcolor{blue!8}\spbase & 28.9 & 56.3 & 11.0 & 13.7\\
\rowcolor{blue!15}\spinstruct & 4.2 & 7.1 & 10.7 & 15.8 \\
\bottomrule
\end{tabular}}
\scriptsize*We were unable to reproduce SpiritLM's ASR performance; therefore, we report their self-reported LS results using ten-shot prompts.\\
\caption{WER on various ASR test sets.}
\label{tab:results_asr}
\end{table}



%% file: latex/sections/4_mt_results.tex
\subsection{MT}\label{sec:texteval}

\input{latex/tables/MT_table}
Having demonstrated that our training approach works well to initially equip \tower{} with speech processing capabilities, we now turn to MT to investigate whether \spire{} can maintain \tower{}'s strong performance on MT despite its speech-centric CPT. 

\paragraph{Datasets and Metrics}
We evaluate on two datasets for MT: FLORES-200~\cite{nllb2024scaling}, which covers \spire{}'s languages, and the WMT23 test set~\citep{kocmi-etal-2023-findings}, which covers \langpb{en}{\{de, ru, zh\}}.
We report COMET-22~\cite[COMET; ][]{rei-etal-2022-comet} and spBLEU\footnote{\texttt{nrefs:1|case:mixed|eff:no|tok:flores200|\\smooth:exp|version:2.5.1}} \cite{papineni-etal-2002-bleu} scores via the SacreBLEU toolkit~\cite{post-2018-call}.

\paragraph{Baselines} We compare the \spire{} models to the text-to-text translation performance of SeamlessM4T.
Additionally, we report the performance of \tower{\textsc{base-7b}} and \tower{\textsc{instruct-7b}}.

\paragraph{Results}
Our results show that even after the speech-centric CPT and mixed speech and text IT stage, the \spire{} models retain the original text-only performance of \tower{} on both FLORES (Table~\ref{tab:flores}) and WMT23 (Table~\ref{tab:wmt}).
This indicates that neither CPT nor IT on speech data negatively impacts 
the model's ability to perform MT.
This is true for both \spbase{}, which achieves performance comparable to \tower{\textsc{base}}; and for IT models, where \spinstruct{} slightly surpasses the performance of \tower{\textsc{instruct}} on \langp{en}{xx}.
Table~\ref{tab:mt-sig} shows that between \tower{\textsc{instruct}} and \spinstruct{}, neither model consistently shows a significant improvement over the other.
\spinstruct{} also outperforms SeamlessM4T by both metrics on all WMT23 language pairs, and for both \langp{en}{xx} and \langp{xx}{en} on FLORES.
\paragraph{Translation-related Tasks}

We follow the evaluation set-up from \tower{}~\citep{alves2024tower} to additionally evaluate \spire{} on translation-related tasks. 
In Table~\ref{tab:related} we report our results on 
APE for \langpb{en}{\{de, ru, zh\}} and NER 
for \{de, en, es, fr, it, pt, zh\}. \spire{} performs similarly to \tower{\textsc{instruct}} across both tasks and all language directions, maintaining the original text-only capabilities even after training on speech data. 
\input{latex/tables/related_tasks}

\input{latex/tables/wmt_table}
\input{latex/tables/mt-significance}

%% file: latex/tables/MT_table.tex
\begin{table}[t]
    \centering
    \resizebox{\columnwidth}{!}{
    \begin{tabular}{lrrrr}
    \toprule
    & \multicolumn{2}{c}{\langp{\textbf{en}}{\textbf{xx}}} & \multicolumn{2}{c}{\langp{\textbf{xx}}{\textbf{en}}} \\
    & C22 & spB & C22 & spB\\
    \midrule
    SeamlessM4T & 87.22 & 39.0 & 87.42 & 39.9\\
    \tower{\textsc{base-7b}} & 87.38 & 37.8 & 88.02 & 41.7\\
    \tower{\textsc{instruct-7b}} & 88.45 & 38.8 & \textbf{88.27} & \textbf{42.0}\\
    \midrule
    \textit{Our models} \\
\hdashline 
    \rowcolor{blue!8}\spbase & 87.41 & 37.4 & 87.97 & 41.4\\
    \rowcolor{blue!15}\spinstruct & \textbf{88.54} & \textbf{39.3} & 88.21 & 41.8 \\
    \bottomrule
    \end{tabular}
    }
    \caption{COMET-22 (C22) and spBLEU (spB) on the FLORES devtest set between English and the other languages supported by \tower{} And \spire{}.}
    \label{tab:flores}
\end{table}


%% file: latex/tables/related_tasks.tex
\begin{table}
    \centering
    \resizebox{\columnwidth}{!}{
    \begin{tabular}{lrrr}
    \toprule
    & \multicolumn{2}{c}{\textbf{APE}} & \multicolumn{1}{c}{\textbf{NER}} \\
    & \langp{en}{xx} & \langp{xx}{en} & Multilingual \\
    \midrule
    \tower{\textsc{Instruct-7b}} & 83.08 & \textbf{80.29} & \textbf{71.56} \\
    \rowcolor{blue!15}\spinstruct &  \textbf{83.13} & 80.08 & 67.10\\
    \bottomrule
    \end{tabular}
    }
    \caption{Results on APE (COMET) and NER (seq. F1).}
    \label{tab:related}
\end{table}

%% file: latex/tables/wmt_table.tex
\begin{table*}[t]
    \centering
    \resizebox{\textwidth}{!}{\begin{tabular}{lrrrrrrrrrrrr}
    \toprule
         & \multicolumn{2}{c}{\langp{\textbf{en}}{\textbf{de}}} & \multicolumn{2}{c}{\langp{\textbf{en}}{\textbf{ru}}} & \multicolumn{2}{c}{\langp{\textbf{en}}{\textbf{zh}}} & \multicolumn{2}{c}{\langp{\textbf{de}}{\textbf{en}}} & \multicolumn{2}{c}{\langp{\textbf{ru}}{\textbf{en}}} & \multicolumn{2}{c}{\langp{\textbf{zh}}{\textbf{en}}}\\
         & C22 & spB & C22 & spB & C22 & spB & C22 & spB & C22 & spB & C22 & spB\\
    \midrule
    SeamlessM4T & 77.76 & 27.8 & 83.22 & 34.2 & 80.14 & 29.7 & 78.69 & 26.6 & 80.58 & 32.5 & 76.96 & 23.8 \\
    \tower{\textsc{base-7b}} & 79.96 & 36.1 & 83.08 & 34.2 & 83.49 & 33.3 & 83.56 & 41.1 & 80.06 & 32.7 & 78.48 & 23.5\\
    \tower{\textsc{instruct-7b}} & 82.34 & 38.8 & \textbf{84.66} & \textbf{34.9} & 85.09 & 35.3 & 84.95 & 45.1 & \textbf{82.94} & \textbf{36.7} & \textbf{80.14} & 26.1\\
        \midrule
            \textit{Our models} \\
\hdashline 
    \rowcolor{blue!8}\spbase & 79.88 & 34.7 & 83.04 & 33.7 & 83.85 & 32.4 & 83.19 & 40.5 & 80.20 & 32.4 & 78.65 & 23.1\\
    \rowcolor{blue!15}\spinstruct & \textbf{82.50} & \textbf{39.5} & 84.60 & \textbf{34.9} & \textbf{85.37} & \textbf{37.3} & \textbf{85.24} & \textbf{45.2} & 82.58 & 36.4 & 79.92 & \textbf{26.3}\\
    
    \bottomrule
    \end{tabular}}
    \caption{COMET-22 (C22) and spBLEU (spB) on the WMT23 test set.}
    \label{tab:wmt}
\end{table*}


%% file: latex/tables/mt-significance.tex
\begin{table}
    \centering
    \resizebox{\columnwidth}{!}{
    \begin{tabular}{llrrr}
    \toprule
    Corpus & Metric & TI Better & SF Better & NS\\%
    \midrule
    FLORES & spBLEU & 3 & 4 & 11 \\
    {} & COMET & 1 & 1 & 16 \\
    \midrule
    WMT23 & spBLEU & 0 & 2 & 4\\
    {} & COMET & 2 & 2 & 2\\ 
    \bottomrule
    \end{tabular}
    }
    \caption{Counts of language pairs in which \tower{\textsc{Instruct}} significantly outperforms \spinstruct{} (TI Better), \spinstruct{} significantly outperforms \tower{\textsc{Instruct}} (SF Better), or differences are not significant (NS). Significance is reported at the $p < 0.5$ level. We used the paired bootstrap method for spBLEU.
    }
    \label{tab:mt-sig}
\end{table}

%% file: latex/sections/4_st_results.tex
\subsection{ST}\label{sec:asteval}

As \spire{} has shown success at both ASR and MT, we now investigate its performance on ST.

\paragraph{Datasets} For ST, we evaluate our models on FLEURS~\cite{conneau2023fleurs}, covering ST between all \langp{en}{xx} pairs, and CoVoST-2~\cite{wang2021covost} for \langp{en}{\{de, zh\}}.

\paragraph{ST approaches} As well as direct ST, we report self-cascades, in which each model transcribes the audio before translating its own output to the target language~(\textit{i.e.}, ASR followed by MT).

\paragraph{Baselines}
We compare \spire{} to SeamlessM4T in both direct and cascaded settings.
We also report the results of SALMONN and Qwen2-Audio, which are both 7B parameter models, like \spire{}.
However, SALMONN and Qwen2-Audio do not support text-to-text translation, so we use them only for direct ST.\footnote{Although Whisper is frequently used for ST, we exclude it because it only supports to-English translation, whereas \spire{} is a from-English ST model. Therefore ST comparisons between the two models are impossible.}
There are also coverage differences between the models: while SeamlessM4T can handle all of \spire{}'s language pairs, neither SALMONN nor Qwen2-Audio supports \langp{en}{ko}; SALMONN also does not support \langp{en}{ru}.

\paragraph{Results}

\input{latex/tables/covost_table}

Our FLEURS spBLEU ST results are reported in Table~\ref{tab:fleurs}. For brevity, 
COMET-22 scores are reported in Appendix \ref{appendix:st_results}.
SeamlessM4T performs best at direct ST for all language pairs except \langp{en}{zh}.
Among the 7B parameter models, \spinstruct{} is the best direct model on average, notably beating SALMONN on all language pairs except \langp{en}{zh}.
It also outperforms Qwen2-Audio on 6 out of 8 language pairs that Qwen2-Audio supports, and ties or beats it for all except \langp{en}{zh} and \langp{en}{de}.

Performance on CoVoST-2 (Table~\ref{tab:covost}) tells a different story.
Although \spinstruct{} maintains its advantage over SeamlessM4T in self-cascaded translation, it attains the worst performance on \langp{en}{zh}, while performing similarly to SALMONN for \langp{en}{de}.
This shows that the direct ST performance of \spinstruct{} is dataset-dependent, which could be a consequence of its relatively small training data. 

\spinstruct{} achieves the best self-cascaded performance by a significant margin for both datasets, outperforming SeamlessM4T by a large margin in this setting.
This demonstrates that \spinstruct{} maintains greater robustness to its own outputs than SeamlessM4T, supporting the insight that LLM-based translation models can be very robust to perturbations \citep{robustmt}.

\input{latex/tables/ST_table.tex}

%% file: latex/tables/covost_table.tex
         

\begin{table}[h]
    \centering
    \resizebox{0.8\columnwidth}{!}{\begin{tabular}{lrrrr}
    \toprule
         & \multicolumn{2}{c}{\langp{en}{de}}  & \multicolumn{2}{c}{\langp{en}{zh}} \\
         & C22 & spB & C22 & spB\\
    \midrule
        \textit{Self-cascade} \\
\hdashline 
    SeamlessM4T & 72.40 & 21.7 & 72.32 & 17.0\\
    \rowcolor{blue!15}\spinstruct & \textbf{78.05} & \textbf{31.8} & \textbf{79.50} & \textbf{28.1}\\
        \midrule
            \textit{Direct} \\
\hdashline
    SALMONN & 74.98 & 22.7 & 80.92 & 27.8\\
    Qwen2-Audio & 82.29 & 34.5 & \textbf{85.27} & \textbf{38.7}\\
    SeamlessM4T & \textbf{85.95} & \textbf{42.3} & 83.62 & 31.3\\
    \rowcolor{blue!15}\spinstruct & 73.96 & 25.4 & 74.53 & 21.0\\
    
    \bottomrule
    \end{tabular}}
    \caption{ST results on CoVoST-2.}
    \label{tab:covost}
\end{table}

%% file: latex/tables/ST_table.tex
\begin{table*}[t]
    \centering

    \resizebox{0.8\textwidth}{!}{\begin{tabular}{lrrrrrrrrrrrr}
    \toprule
        & de & es & fr & it & ko & nl & pt & ru & zh & $\text{avg}_\text{7}$ & $\text{avg}_\text{all}$ \\

         \midrule
        \multicolumn{11}{l}{\textit{Self-Cascade}} \\
        \hdashline 
        SeamlessM4T & 24.2 & 21.5 & 37.7 & 18.9 & 12.5 & 16.9 & 28.2 & 27.1 & 14.6 & 23.1 &22.4\\
        \spinstruct & \textbf{38.1} & \textbf{29.4} & \textbf{45.3} & \textbf{31.2} & \textbf{23.1} & \textbf{31.2} & \textbf{42.9} & \textbf{33.5} & \textbf{29.0} & \textbf{35.3} &\textbf{33.7}\\
        \midrule
        \multicolumn{11}{l}{\textit{Direct}} \\
        \hdashline 
         SeamlessM4T & \textbf{39.2} & \textbf{28.0} & \textbf{48.1} & \textbf{30.6} & \textbf{21.5} & \textbf{30.8} & \textbf{47.5} & \textbf{34.3} & 23.2 & \textbf{35.3} & \textbf{33.7}\\
        SALMONN & 25.5 & 20.8 & 34.3 & 16.7 & 0.1 & 20.5 & 32.6 & 3.1 & 21.9 & 24.6  &19.5\\
        Qwen2-Audio & 31.8 & 23.5 & 31.3 & 23.5 & 5.4 & 22.3 & 36.1 & 23.7 & \textbf{24.7}& 27.6 &24.7\\
         \rowcolor{blue!15}\spinstruct & 31.1 & 23.5 & 37.9 & 25.5 & 15.4 & 25.7 & 37.3 & 26.9 & 21.0 & 28.9 & 27.1\\
    \bottomrule
    \end{tabular}}
    \caption{FLEURS ST \langp{ex}{xx} results with self-cascade and direct models in terms of spBLEU. The $\text{avg}_\text{7}$ column averages over the 7 language pairs that all models in the table support (excluding \langp{en}{\{ko, ru\}}).}
    \label{tab:fleurs}
\end{table*}

%% file: latex/sections/6_conclusion.tex
\section{Analysis}
The key innovation of our approach is the application of the CPT followed by IT paradigm to discretized speech allowing us to build upon existing text-only capabilities of our base model. Here, we analyze how the composition of these two training phases contributes overall to model performance across all tasks previously evaluated.
To that end, we consider several variants of \spbase{} and \spinstruct{} which are described in Table~\ref{tab:our_models} and whose results are reported in Table \ref{tab:analusis}.
\begin{itemize}
    \item \textit{i}) No CPT was performed and IT was performed with the entire IT data mix (\itonly);
    \item \textit{ii}) CPT was performed and no data from \textsc{TowerBlocks} was seen during IT (\noblocks{}), and
    \item \textit{iii}) CPT was performed and pseudo-labeled ST data and FLEURS were omitted from the IT data mix (\nopseudo{}).
\end{itemize}
We report additional datasets in Appendix \ref{appendix:ablation}.
\input{latex/tables/our_models}

\paragraph{Effectiveness of CPT and IT} Our previous results demonstrated that using both CPT and IT was the most effective strategy. The performance gap between \spinstruct{} and the \itonly{} on ASR~(5.3 points in LS test-clean) further shows that IT alone is also not as effective. However, for ST we observe that performing only IT leads to a strong model that is capable of performing speech translation.
This contrasts from \spbase{}, for which we also attempted direct ST but the model failed to produce output in the target language, even when given few-shot prompts. Despite the impressive results from \itonly{}, we still observe the best performance by \spinstruct{} showing that while the effect of CPT is not as drastic as in the case of ASR, we still observe gains with a speech-centric CPT phase.

\paragraph{Modality Interplay} Our results show that text and speech modalities are orthogonal to each other. Specifically, the performances of \itonly{} and \spinstruct{} show that speech-centric CPT \textit{does not} degrade the text performance of the base model. However, MT quality suffers when \tower{\textsc{Blocks}} is removed from the IT data, as is shown by \noblocks{}'s much weaker performance than \spinstruct{}. Simultaneously, \spinstruct{} performs on par with \noblocks{} on both ASR and ST, indicating that adding text instructions also \textit{does not} degrade performance on speech tasks. It is worth highlighting that 
a model strong at both MT and ASR~(\nopseudo{}) does not lead to a strong ST model, showing surprisingly that competence at MT is not very helpful for direct ST.

\input{latex/tables/analysis_table}

\section{Conclusion}

In this work we presented \spire{}, a simple and effective recipe for adapting a text-based, translation-specialist LLM to the speech modality 
while preserving the original performance on text-based tasks. We investigated the impact of speech integration on two stages of LLM adaptation, CPT and IT, finding that both contribute to the final model's performance on speech tasks.  
Our results demonstrate that we are able to successfully integrate a new modality without compromising the original model's capabilities. \spire{} achieves competitive performance on ASR, while its MT abilities remain on par with the original \tower{} model. Finally, for the ST task, we find that the leveraging ASR and MT data does not directly transfer to ST performance. Nonetheless, the model achieves promising performance with both direct and self-cascaded ST. To benefit the community, we only used publicly available and licensed data to train our models, making our results reproducible.
As future work, we intend to extend this recipe to multilingual settings by replacing our English HuBERT speech component by the multilingual mHuBERT-147~\cite{boito2024mhubert}.

\section*{Limitations}
The downstream tasks we evaluate on are restricted to MT and ASR/ST, which provides an idea of the model performance but do not give us the full picture. 
We plan to address this by utilizing the LM-harness evaluation~\cite{eval-harness} to evaluate on a suite of text-based benchmarks such as MMLU~(multitask language understanding)~\cite{hendrycks2021ethics,hendryckstest2021}, Arc~(commonsense reasoning) \cite{allenai:arc}, Belebele~(reading comprehension)~\cite{bandarkar-etal-2024-belebele}, and HellaSwag~(sentence completion) \cite{zellers-etal-2019-hellaswag}. Lastly, our model handles speech and text on the input side but is currently limited to generating only text. 

\section*{Acknowledgments}
This work was supported by EU's Horizon Europe Research and Innovation Actions (UTTER, contract 101070631), by UK Research and Innovation (UKRI) under the UK government’s Horizon Europe funding guarantee (grant number 10039436: UTTER), by the project DECOLLAGE (ERC-2022-CoG 101088763), by the Portuguese Recovery and Resilience Plan through project C645008882-00000055 (Center for Responsible AI), by Fundação para a Ciência e Tecnologia (FCT) through the project with reference UIDB/50021/2020 (DOI:10.54499/UIDB/50021/2020), and by FCT/MECI through national funds and when applicable co-funded EU funds under UID/50008: Instituto de Telecomunicações.
This work was performed using HPC resources from GENCI–IDRIS (Grant 2023-AD011014668R1).
We thank Duarte Alves and Giuseppe Attanasio for their insightful comments.

%% file: latex/tables/our_models.tex
\begin{table}[t]
    \centering
    \resizebox{\columnwidth}{!}{\begin{tabular}{cc|cc|ccc}
        \toprule
         \multirow{2}{*}{Model} & \multirow{2}{*}{Base Model} & \multicolumn{2}{c|}{CPT} & \multicolumn{3}{c}{IT} \\
          &  & Speech & Text & Speech & Pseudo & Text \\
         \midrule
        \itonly{} & TowerBase-7B & \ding{55} & \ding{55} & \ding{51} & \ding{51} & \ding{51} \\
        \rowcolor{blue!8}\spbase{} & SpireBase & \ding{51} & \ding{51} & \ding{55} & \ding{55} & \ding{55} \\
        \rowcolor{blue!15}\spinstruct{} & SpireBase & \ding{51} & \ding{51} & \ding{51} & \ding{51} & \ding{51} \\ 
        \hdashline
        \multicolumn{7}{c}{\textit{\spire{} Variants}}\\
        \hdashline
        \noblocks{} & SpireBase & \ding{51} & \ding{51} & \ding{51} & \ding{51} & \ding{55}\\
        \nopseudo{} & SpireBase & \ding{51} & \ding{51} & \ding{51} & \ding{55} & \ding{51}\\
        \bottomrule
    \end{tabular}}
    \caption{Ablations of our models. The CPT and IT columns indicate which data was seen during training.}
    \label{tab:our_models}
\end{table}

%% file: latex/tables/analysis_table.tex
\begin{table}[t]
    \centering
\resizebox{\columnwidth}{!}{\begin{tabular}{lrrrrrrr}
    \toprule
        & \multicolumn{1}{c}{\textbf{ASR}} & \multicolumn{4}{c}{\textbf{MT}} & \multicolumn{2}{c}{\textbf{ST}} \\
        \midrule
        &  & \multicolumn{2}{c}{\langp{en}{xx}} &  \multicolumn{2}{c}{\langp{xx}{en}}  & \multicolumn{2}{c}{\langp{en}{xx}}\\
        \midrule
        &\multicolumn{1}{c}{WER}& C22 & spB & C22 & spB &  C22 & spB \\
        \midrule
    \rowcolor{blue!15}\spinstruct{} & 4.2 & 88.54 & 39.3 & 88.21 & 41.8 & \textbf{81.33} & \textbf{27.1}\\
    \hdashline
    \itonly  & 9.5  & \textbf{88.57} & \textbf{39.4} & 88.17 & 41.7 & 79.10 & 26.1\\
    \noblocks & 4.1 & 82.98 & 34.2 & 85.93 & 36.1 & 81.11 & 27.1\\
    \nopseudo &\textbf{3.9} & 88.40 & 38.9 & \textbf{88.22} & \textbf{42.0} & 62.80 & 27.1\\
    \bottomrule
    \end{tabular}}
    \caption{Ablation models and \spinstruct{} on LS Clean for ASR, FLORES devtest for MT, and Fleurs for ST reporting WER, COMET-22 (C22), and spBLEU (spB).}
    \label{tab:analusis}
\end{table}

%% file: latex/sections/6_appendix.tex
\clearpage

\appendix

\label{sec:appendix}
\section{Data}
\subsection{Speech Data Preprocessing}\label{appendix:datanorm}

\paragraph{Normalization} In order to make transcripts consistent across the different datasets, the following normalization is applied:

\begin{itemize}
    \item \textbf{GigaSpeech (CPT):} we lower-case the text and replace punctuation tags: \texttt{<COMMA>}, \texttt{<PERIOD>}, \texttt{QUESTIONMARK>}, \texttt{<EXCLAMATIONPOINT>} with their appropriate punctuation.
    \item \textbf{MLS (CPT):} we apply a tail-end normalization step here which uniformly samples each speaker to have at maximum 13 transcriptions. This allows us to have a better distribution of speakers.
    \item \textbf{CV (IT):} we subsampled from CommonVoice to ensure a minimum duration of 3 seconds per sample. To enhance transcript diversity, we limit each transcript to 4 unique speakers.
\end{itemize}

\paragraph{Deduplication} As in previous work~\cite{zhang-etal-2023-speechgpt,rubenstein2023audiopalm,chang2024exploring}, we merge consecutive repeated DSU tokens into a single token to reduce sequence length.

\subsection{Prompt Format}\label{appendix:prompts}
Table \ref{tab:prompts} show the prompts used during both training stages. 
\input{latex/tables/prompts_main}

\section{CTC-based ASR model}\label{appendix:model:asrmodel}

We train a CTC-based ASR model using the HuggingFace \texttt{transformers} library~\cite{wolf2019huggingface}, leveraging the ASR data from the IT stage~(CV, Table~\ref{tab:speechstatistics}) as training data after whisper normalization. Our ASR model is made of the HuBERT-Large\footnote{\scriptsize\url{https://huggingface.co/facebook/hubert-large-ll60k}} speech representation model, followed by three hidden layers and a vocabulary projection layer. We train for 50 epochs with a dropout of 0.3 and a learning rate of 1e-4 with a warm-up ratio of 0.15. We perform step-based best checkpoint selection based on CER scores.
Our best checkpoint was obtained at step 220K~(at epoch 12.8).

\section{ST results}\label{appendix:st_results}

Table \ref{tab:fleurs_c} report results of ST on FLEURS across baseline models and \spinstruct{}. We report COMET-22. We observe the same trend in scores as reported by spBLEU where in \spinstruct{} obtains the best self-cascaded performance while beating Qwen2-Audio and SALMONN on direct ST across most language pairs. SeamlessM4T obtains the overall best performance in direct ST.

\begin{table*}[t]
    \centering

    \resizebox{\textwidth}{!}{\begin{tabular}{lrrrrrrrrrrrr}
    \toprule
        & de & es & fr & it & ko & nl & pt & ru & zh & $\text{avg}_\text{7}$ & $\text{avg}_\text{all}$ \\
         \midrule
        \multicolumn{11}{l}{Self-Cascade} \\
        \hdashline 
        SeamlessM4T & 72.69 & 76.97 & 78.06 & 76.03 & 75.33 & 72.58 & 78.25 & 79.38 & 69.76 & 74.91 & 75.45\\
        \rowcolor{blue!15}\spinstruct & \textbf{84.26} & \textbf{83.32} & \textbf{84.70} & \textbf{85.16} & \textbf{86.89} & \textbf{84.91} & \textbf{86.01} & \textbf{86.45} & \textbf{85.21} & \textbf{84.80} & \textbf{85.21}\\
        \midrule
        \multicolumn{11}{l}{Direct} \\
        \hdashline 
         SeamlessM4T & \textbf{84.79} & \textbf{83.20} & \textbf{85.32} & \textbf{85.03} & \textbf{85.17} & \textbf{85.17} & \textbf{86.75} & \textbf{86.31} & 79.90 & \textbf{84.31} & \textbf{84.63}\\
        SALMONN & 77.41 & 77.99 & 79.95 & 74.47 & 61.07 & 77.18 & 80.94 & 53.05 & \textbf{81.63} & 78.51 &73.74\\
        Qwen2-Audio & 79.82 & 80.43 & 79.44 & 81.28 & 69.33 & 78.75 & 83.41 & 77.90 & 80.71 & 80.55  &79.01\\
         \rowcolor{blue!15}\spinstruct & 80.16 & 79.82 & 80.68 & 81.63 & 82.62 & 81.93 & 83.18 & 82.19& 79.76 & 81.02 & 81.33\\

    \bottomrule
    \end{tabular}}
    \caption{FLEURS ST \langp{ex}{xx} results with self-cascade and direct models in terms of COMET-22. $\text{avg}_\text{7}$ covers the 7 language pairs that all models in the table support (excluding \langp{en}{\{ko, ru\}}).}
    \label{tab:fleurs_c}
\end{table*}

\section{Ablation results}\label{appendix:ablation}

Table \ref{tab:asr_ablation} reports results from all remaining evaluation datasets across ASR, MT, and ST. We report the same metrics as in Section \ref{sec:experiments}. Here as well, we note that in MT, the inclusion of speech data did not degrade text-only performance (\spinstruct{} vs. \itonly). Similarly, the inclusion of task-specific text data also did not harm performance on ASR (\noblocks{} vs. \spinstruct{}). Lastly, \spinstruct{} has the best performing direct ST system, further showing that individual task competencies (in MT and ASR) do not contribute directly to a compositional task (ST) but rather the inclusion of task-specific data leads to the highest gains (\nopseudo{} vs \spinstruct{}).

\begin{table*}[t]
    \centering
\resizebox{\textwidth}{!}{\begin{tabular}{lrrrrrrrrr}
    \toprule
        & \multicolumn{3}{c}{\textbf{ASR}} & \multicolumn{4}{c}{\textbf{MT}} & \multicolumn{2}{c}{\textbf{ST}} \\
        \midrule
        &\multicolumn{3}{c}{WER}& C22 & spB & C22 & spB &  C22 & spB \\
        \midrule
        & LS Other & Fleurs & VoxPopuli  & \multicolumn{2}{c}{\langp{en}{xx}} &  \multicolumn{2}{c}{\langp{xx}{en}}  & \multicolumn{2}{c}{\langp{en}{xx}}\\
    \rowcolor{blue!15}\spinstruct{} & \textbf{7.1} & 10.7 & 15.8 & 84.16 & \textbf{37.2} & \textbf{82.58} & \textbf{41.8} & \textbf{81.33} & \textbf{27.1}\\
    \hdashline
    \itonly  & 13.8 & 14.3 & 40.7 & \textbf{84.19} & 36.9 & 82.25 & 35.6 & 71.52 & 20.1\\
    \noblocks& 7.4 & \textbf{10.4} & 15.8 & 73.12 & 26.9 & 74.78 & 25.1 & 74.02 & 23.2\\
    \nopseudo & 7.3 & 11.1 & \textbf{14.3} & 83.93 & 36.9 & 82.50 & 35.9 & 59.88 & 6.8\\
    \bottomrule
    \end{tabular}}
    \caption{Ablation models and \spinstruct{} on LS Other, Fleur, VoxPopuli for ASR, WMT23 for MT, and CoVoST-2 for ST reporting WER, COMET-22 (C22), and spBLEU (spB).}
    \label{tab:asr_ablation}
\end{table*}

%% file: latex/tables/prompts_main.tex
\begin{table}[h]
\centering\small
\begin{tabular}{l}
\toprule
\multicolumn{1}{c}{\textbf{ASR (CPT)}} \\ 
\midrule
\begin{tabular}[c]{@{}l@{}}
\texttt{Speech:\textless extra\_id\_i\textgreater$\cdots$\textless extra\_id\_j\textgreater} \\
\texttt{English: \{TRANSCRIPT\}}
\end{tabular} \\ 
\midrule
\multicolumn{1}{c}{\textbf{MT (CPT)}} \\
\midrule
\begin{tabular}[c]{@{}l@{}}
\texttt{Source\_lang: Source-sentence} \\
\texttt{Target\_lang: \{TRANSLATION\}}
\end{tabular} \\ 
\midrule
\multicolumn{1}{c}{\textbf{ASR (IT)}} \\ 
\midrule
\begin{tabular}[c]{@{}l@{}}
\texttt{Speech: \textless extra\_id\_i\textgreater$\cdots$\textless extra\_id\_j\textgreater} \\
\texttt{English: \texttt{\{TRANSCRIPT\}}} 
\end{tabular} \\ 
\midrule
\multicolumn{1}{c}{\textbf{Direct ST (IT)}} \\ 
\midrule
\begin{tabular}[c]{@{}l@{}}
\texttt{Speech: \textless extra\_id\_i\textgreater$\cdots$\textless extra\_id\_j\textgreater} \\
\texttt{TARGET\_LANG: \texttt{\{TRANSLATION\}}}
\end{tabular} \\ 
\midrule
\multicolumn{1}{c}{\textbf{Multi-turn ST (IT)}} \\ 
\midrule
\begin{tabular}[c]{@{}l@{}}
\texttt{Speech: \textless extra\_id\_i\textgreater$\cdots$\textless extra\_id\_j\textgreater} \\
\texttt{English:\texttt{\{TRANSCRIPT\}}} \\
\texttt{TARGET\_LANG: \texttt{\{TRANSLATION\}}}
\end{tabular} \\
\bottomrule
\end{tabular}
\caption{Prompt formats for CPT and IT.}
\label{tab:prompts}
\end{table}